\documentclass[11pt]{article}

\PassOptionsToPackage{table}{xcolor}
\usepackage[preprint]{acl}

\usepackage{times}
\usepackage{latexsym}
\usepackage[T1]{fontenc}
\usepackage[utf8]{inputenc}
\usepackage{microtype}
\usepackage{inconsolata}

\usepackage{amsmath,amssymb}
\usepackage{textcomp}
\usepackage{graphicx}
\usepackage{booktabs}
\usepackage{multirow}
\usepackage{array}
\usepackage{tabularx}
\usepackage{colortbl}
\usepackage{capt-of}
\usepackage{placeins}

\usepackage{enumitem}
\usepackage{fvextra}
\usepackage[most]{tcolorbox}
\usepackage{listings}
\usepackage{xcolor}

\graphicspath{{figures/}}

\definecolor{PromptGray}{RGB}{105,105,105}
\definecolor{PromptBackground}{RGB}{248,248,248}

\newtcblisting{PromptBlock}[1]{
  enhanced,
  breakable,
  listing only,
  colback=PromptBackground,
  colframe=PromptGray,
  colbacktitle=PromptGray,
  coltitle=white,
  title={#1},
  fonttitle=\bfseries,
  boxrule=0.8pt,
  arc=1.2mm,
  outer arc=1.2mm,
  left=2mm,
  right=2mm,
  top=1.5mm,
  bottom=1.5mm,
  toptitle=1.5mm,
  bottomtitle=1.5mm,
  before skip=4pt,
  after skip=4pt,
  listing options={
    basicstyle=\ttfamily\footnotesize,
    columns=fullflexible,
    keepspaces=true,
    breaklines=true,
    showstringspaces=false,
    literate={"}{{\char34}}1
  }
}

\title{SimRec: Learning to Search, Ask, and Recommend from Simulated Interaction}

\date{}
\author{
\textbf{Yaxin Zhang}\textsuperscript{1},
\textbf{Xingyu Wu}\textsuperscript{1},
\textbf{Yuchen Yan}\textsuperscript{1},
\textbf{Yuqin Dai}\textsuperscript{2},
\textbf{Hong Zhao}\textsuperscript{3} \\
\textbf{Lixing Shen}\textsuperscript{3},
\textbf{Zhan Chen}\textsuperscript{3},
\textbf{Jian Shao}\textsuperscript{1},
\textbf{Yongliang Shen}\textsuperscript{1} \\[0.6em]
\mbox{%
\textsuperscript{1}Zhejiang University \quad
\textsuperscript{2}Tsinghua University \quad
\textsuperscript{3}Hikvision Research Institute%
}\\
\texttt{\{zyx221, syl\}@zju.edu.cn}
}

\newcommand{\abldrop}[1]{{\color{red!75!black}\scriptsize$(-#1)$}}

\begin{document}
\maketitle
\begin{abstract}
LLM-based recommendation agents can search catalogs, inspect products, ask users, and revise rejected recommendations. However, these behaviors are often governed by hand-designed workflows, while existing training methods optimize isolated components such as query rewriting or list ranking rather than the interaction policy as a whole. We present \textbf{SimRec}, a reinforcement learning framework that trains the full interaction policy of a tool-integrated recommendation agent. SimRec introduces two components. A task-anchored user simulator provides feedback on every recommendation: conditioned on the current task need rather than on user profile, and it can be an external LLM or the recommender itself under self-play. A task-aligned reward optimizes the whole trajectory with GRPO, crediting natural retrieval quality and valid tool use alongside the final hit, with training-only target injection to handle sparse hits over large catalogs. On ESCI and Amazon C4, SimRec improves the NDCG@100 of a Qwen3-4B agent from \textbf{32.23\% to 35.42\%} and from \textbf{22.94\% to 24.72\%}, outperforming training-free agents on far larger backbones and task-aligned baselines on NDCG. Analysis shows that the trained agent verifies candidates, uses preferences, and recovers from rejected recommendations far more often than its untrained counterpart. Our code is available at \url{https://github.com/ZJU-REAL/SimRec/}.

\end{abstract}

\begin{figure}[!htbp]
    \centering
    \includegraphics[width=\columnwidth]{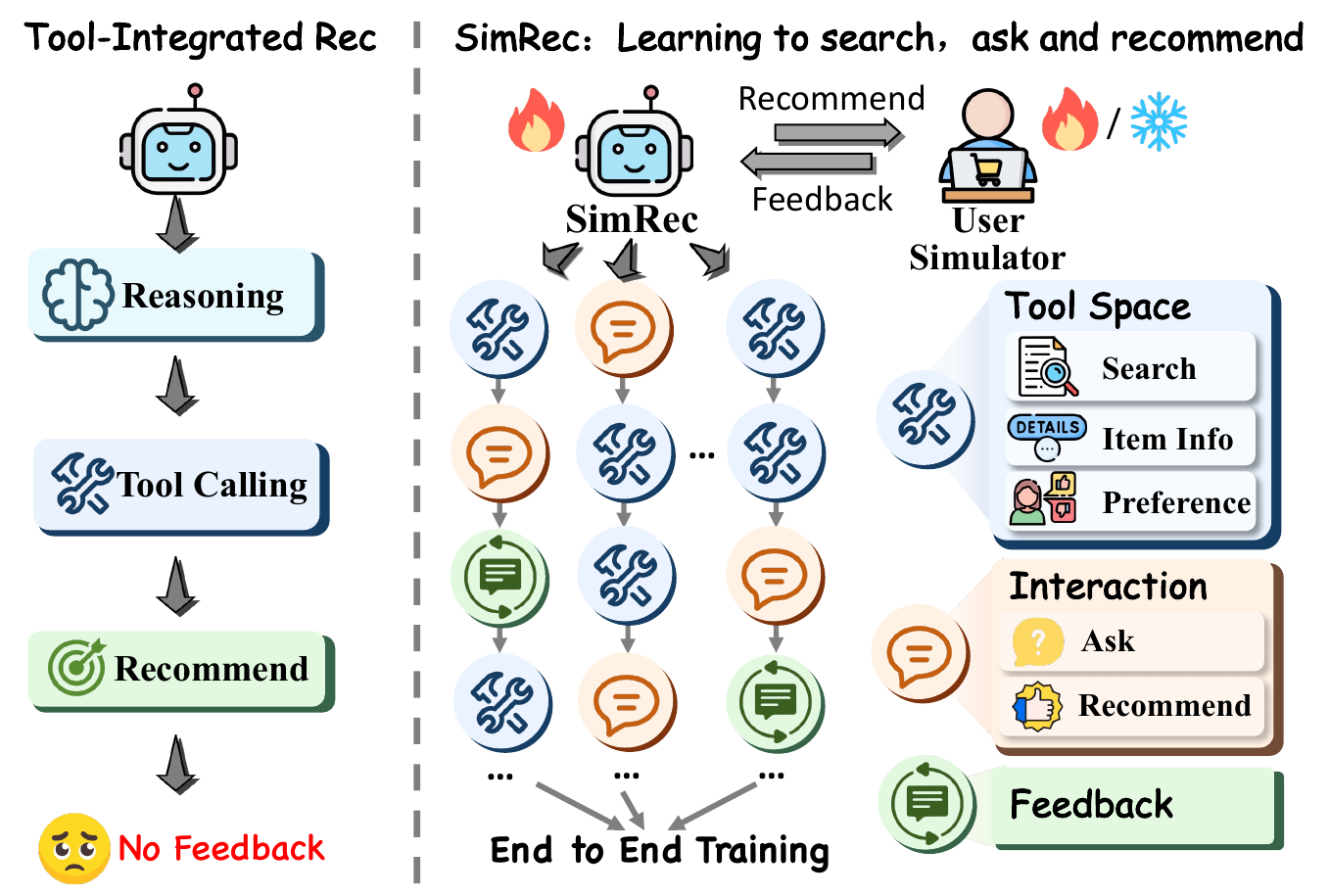}
    \caption{Compared to existing tool-integrated recommender agents (left),
    SimRec (right) learns to search, ask, and recommend from simulated interaction.}
    \label{fig:simrec-motivation}
\end{figure}

\section{Introduction}

\begin{figure}[!htbp]
    \centering
    \includegraphics[width=\columnwidth]{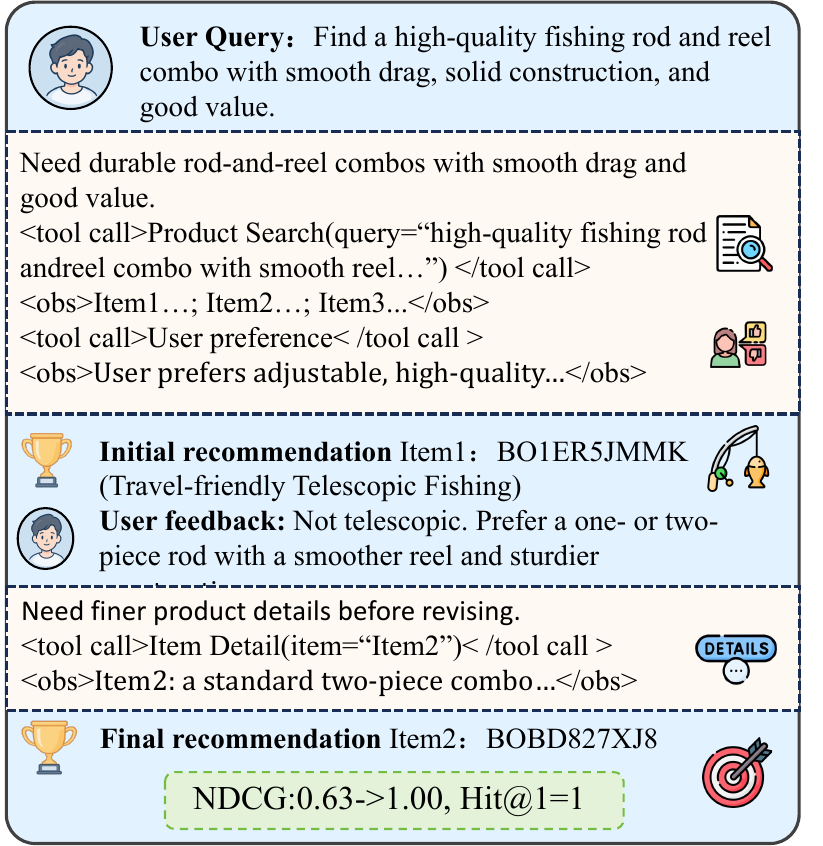}
    \caption{Case study of SimRec multi-turn recommendation refinement. SimRec uses user feedback to revise its recommendation.}
    \label{fig:simrec-case}
\end{figure}

Recommender systems are moving from one-shot prediction toward interaction. Built on a large language model (LLM), a recommendation agent takes a request in natural language, retrieves candidates from the catalog, inspects those that look plausible, asks the user when the need is underspecified, and recommends an item; if the user is not satisfied, it continues from the user's reaction~\citep{huangRecommenderAIAgent2025,zhaoLetMeIt2024b,zhangRecThinkerAgenticFramework2026a,liAgenticRecRecommendationOrientedAgentic2026b,linAutonomousInformationSeeking2026a,wangSelfDistilledReinforcementLearning2026a}. Recommendation quality in this setting is not the output of a single scoring model but the result of a sequence of decisions made over the course of the interaction. A feedback-driven revision is illustrated in Figure~\ref{fig:simrec-case}.

Existing agents make these decisions by rules set in advance rather than by learning from outcomes: a hand-designed workflow prescribes when to search, when to ask, and when to recommend~\citep{huangRecommenderAIAgent2025,fang2024macrs,wang2025grsu}. Where training is used at all, it targets one step lifted out of the interaction, such as rewriting a query for the retriever~\citep{linRecR1BridgingGenerative2026a,yoonAskOptimalQuestions2025,nguyenMiniELMLightweightAdaptive2025}, ranking a given list~\citep{zhuRankgrpoTrainingLlmbased2026,fengExpectationConfirmationPreference2025}, or generating a response to a given dialogue~\citep{xiangUserSimulatorGuidedMultiTurn2026a,caiMGFRecReinforcedReasoning2026}; the outcome of the whole interaction never reaches the policy that made the decisions along the way. These limitations show in behavior. 
Such agents seldom verify the candidates they retrieve, seldom consult the user's history, and rarely recover once a recommendation is rejected, and a larger backbone does not make up for it (Section~\ref{sec:analysis}).
An interaction policy has to be trained on the interaction itself.

Reinforcement learning over complete trajectories trains a policy on exactly this: the agent acts, the environment responds, and the whole trajectory is optimized for the final outcome. For search agents the recipe is well established, with the model interleaving reasoning and search calls and the trajectory rewarded by the correctness of the final answer~\citep{jin2025searchr1,qian2025toolrl,feng2025gigpo}. The search setting, however, comes with its environment given: the index answers every query the same way, search is the only tool, and the answer is either right or wrong (Figure~\ref{fig:simrec-motivation}). Recommendation has none of these. The environment is a user who reacts to each recommended item, the evidence comes from several tools with different costs, and the outcome is a hit that is rare and does not say where the failure occurred. Training a recommendation agent on the interaction itself therefore requires three things to be supplied: the user, the actions, and the objective.
 
Interaction with real users cannot supply training signal at the scale reinforcement learning needs, so the user is played by a simulator~\citep{xiangUserSimulatorGuidedMultiTurn2026a}. A simulator that knows the target item leaks it into the dialogue~\citep{zhu2024reliable,kimStopPlayingGuessing2025}, and one that reconstructs a full user profile may drift from the actual need over turns~\citep{zhang2024agent4rec,meshiConvapparelBenchmarkDataset2026}; the simulator should say only what a real user would say after a recommendation, namely whether the item fits the current need and what it lacks~\citep{biConversationalProductSearch2019}. The actions have to cover the evidence a recommendation depends on, that is, the candidates, the details of a single item, and the user's long-term preferences; each has a cost and none is always needed, so when to draw on each is left to the policy~\citep{liChatcrsIncorporatingExternal2025,maAskRecommendEmpirical2024}. The objective has to credit what the hit cannot see. A hit is rare, and a miss does not say whether the target was never retrieved or was retrieved and not chosen; retrieval quality and tool use therefore have to be rewarded alongside the final hit, and the reward has to remain informative when the hit is absent~\citep{wangSelfDistilledReinforcementLearning2026a}.

To meet these requirements, we propose SimRec, a framework that trains a recommendation agent end-to-end on multi-turn interaction with a simulated user (Figure~\ref{fig:simrec-case}). The user is a task-anchored simulator: it is conditioned on the current task need rather than on the target item or a profile, answers each recommendation with what the item lacks, and never names the target; it can be an external LLM or the recommender itself under shared-parameter self-play. The actions are retrieval, item inspection, preference access, a clarifying question, and a recommendation, in no fixed order; long-term preferences are thus evidence the agent fetches, not part of the simulator. The objective is a task-aligned reward over the whole trajectory that credits natural retrieval quality and valid tool use alongside the final hit, with training-only target injection so that candidate comparison can still be learned when retrieval misses. Trained with GRPO under this objective, the agent learns when to search, when to inspect, when to ask, and when to recommend.

On ESCI and Amazon C4, SimRec raises the NDCG@100 of its Qwen3-4B backbone from 32.23\% to 35.42\% and from 22.94\% to 24.72\%, above training-free agents on far larger backbones and above task-aligned baselines. The gain is in the interaction: the trained agent verifies candidates and consults preferences where the untrained one almost never does, recovers from a rejected recommendation, and ranks a fixed candidate set better in a controlled reranking test. The gains carry over to a sparse retriever and a smaller backbone, and training leaves general capabilities intact.

Our contributions are summarized as follows:
\begin{itemize}[leftmargin=*,topsep=2pt,itemsep=1pt,parsep=0pt,partopsep=0pt]
    \item We treat the interaction decisions of a recommendation agent, namely when to search, inspect, ask, and recommend, as a policy to be trained on the outcome of the whole interaction, and identify what such training requires: the user, the actions, and the objective.
    \item We propose SimRec, which addresses these requirements with a task-anchored user simulator, a tool-integrated action space over retrieval, item inspection, and preferences, and a task-aligned trajectory reward with training-only target injection.
    \item SimRec outperforms retrieval, training-free, and task-aligned baselines on ESCI and Amazon C4, and its gains trace to learned interaction behavior: candidate verification, use of preferences, recovery after a rejection, and stronger candidate comparison.
\end{itemize}

\section{Related Work}

\subsection{Agentic Recommendation Agents}
Conversational recommender systems use natural-language interaction to elicit user needs and support preference understanding and recommendation \cite{christakopoulouConversationalRecommenderSystems2016,jannachSurveyConversationalRecommender2022}. LLM-based methods further improve intent understanding, candidate analysis, response generation, and multi-turn planning \cite{huangSurveyFoundationModelPowered2025,zhangSurveyLargeLanguage2026,liChatcrsIncorporatingExternal2025,duSAPIENTMasteringMultiturn2025}, while conversational product search studies negative feedback, proactive questioning, query reformulation, and preference clarification \cite{biConversationalProductSearch2019,maAskRecommendEmpirical2024,yeProductagentBenchmarkingConversational2025,zouPSConConversationalProduct2025,liWizardShoppingTargetoriented2025}. Recent recommender agents formulate recommendation as sequential observation, reasoning, and action, often augmented with retrieval, user profiles, and external knowledge \cite{huangRecommenderAIAgent2025,zhangRecThinkerAgenticFramework2026a,shangAgentrecbenchBenchmarkingLlm2026,shiPersonaXRecommendationAgentoriented2025,huiSafeHumanalignedGame2026}. Related work also explores proactive retrieval, query reformulation, and task-specific post-training for recommendation and tool decisions \cite{zhaoLetMeIt2024b,linAutonomousInformationSeeking2026a,linRecR1BridgingGenerative2026a,zhuRankgrpoTrainingLlmbased2026,liAgenticRecRecommendationOrientedAgentic2026b}. However, existing multi-turn interaction is still largely used for need elicitation and information acquisition before recommendation. SimRec instead models a complete recommendation loop, where user responses and external evidence continuously update subsequent recommendation decisions.

\subsection{User Simulation for Recommendation}

User simulation has long supported conversational recommender systems (CRSs). Early work developed preference-aware and agenda-based simulators, while UserSimCRS incorporated persona, context, satisfaction, and conditional response generation \cite{zhangEvaluatingConversationalRecommender2020a,afzaliUserSimCRSUserSimulation2023}. Subsequent studies systematize user simulation across information-access tasks and distinguish its training and evaluation roles \cite{balogUserSimulationEvaluating2023,bernardFormalCharacterizationUser2024}.

Recent work increasingly uses LLMs to simulate user profiles, preferences, and conversational behavior \cite{niSurveyLLMbasedConversational2026,wangKnowYouFirst2025,shiPersonaXRecommendationAgentoriented2025,bougieSimuserSimulatingUser2025,zhangLlmpoweredUserSimulator2025}, together with dedicated evaluation protocols and benchmarks \cite{yoonEvaluatingLargeLanguage2024,kimStopPlayingGuessing2025,meshiConvapparelBenchmarkDataset2026}. Simulated interactions have also been used to optimize multi-turn recommendation policies through preference optimization, self-distillation, and ranking-based objectives \cite{xiangUserSimulatorGuidedMultiTurn2026a,wangSelfDistilledReinforcementLearning2026a,fengExpectationConfirmationPreference2025,zhuRankgrpoTrainingLlmbased2026}. In contrast, SimRec anchors simulation to the current task need and generates feedback on recommendation mismatch, while exposing long-term preferences separately through a tool.

\section{Method}
\label{sec:method}
\subsection{Problem Formulation}
\label{sec:problem-formulation}
\begin{figure*}[t]
    \centering
    \includegraphics[width=0.94\textwidth]{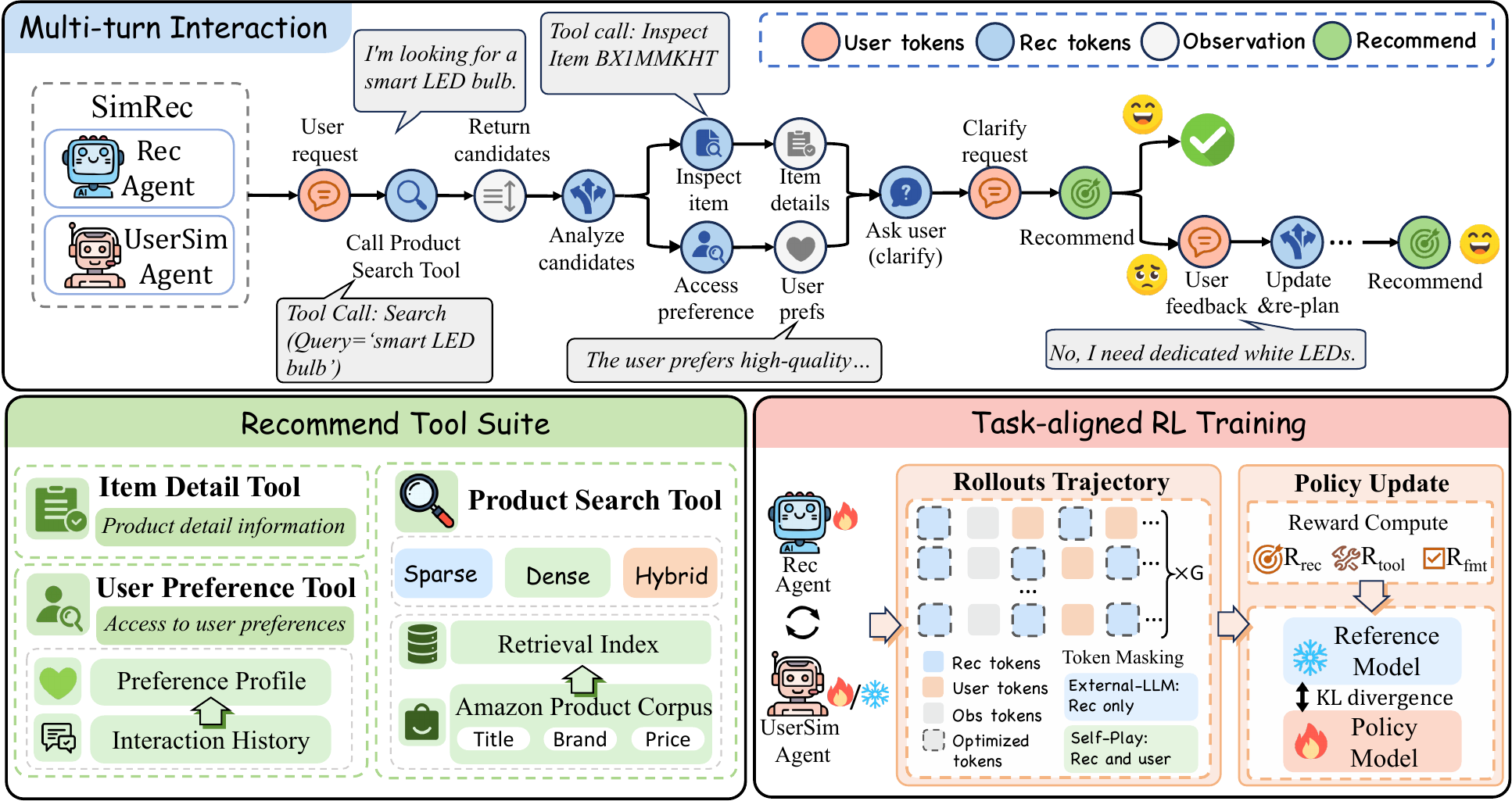}
    \caption{Overview of SimRec. The recommendation agent interacts with a user simulator
    and external recommendation tools, while task-aligned GRPO optimizes the policy over
    complete multi-turn trajectories.}
    \label{fig:simrec-overview}
\end{figure*}

SimRec models recommendation as a finite-horizon interaction among a recommendation agent, external tools, and a user simulator (Figure~\ref{fig:simrec-overview}). Let $\mathcal I$ be the product catalog, whose items have identifiers, textual descriptions, and structured attributes. A task is defined as
\begin{equation}
x=(u_0,h_x,z_x,\mathcal Y_x),
\label{eq:task-definition}
\end{equation}
where $u_0$ is the initial request, $h_x$ is the user's historical interactions,  $z_x$ is the current task need, and $\mathcal Y_x\subseteq\mathcal I$ is the set of target items. The simulator is conditioned on $z_x$, and the environment uses $\mathcal Y_x$ to assess success; neither is directly exposed to the agent. The agent can access long-term preferences summarized from $h_x$ through the preference tool. When training requires one target, $y_x\in\mathcal Y_x$ denotes the selected item; for single-target tasks, $|\mathcal Y_x|=1$.

At step $t$, the state $s_t$ contains the initial request and accumulated dialogue, including user responses and tool observations. The agent selects $a_t\sim\pi_\theta(\cdot\mid s_t)$ from clarification, tool-use, and recommendation actions. Tools support product retrieval, item inspection, and preference access. The interaction follows
\begin{equation}
s_t \rightarrow a_t \rightarrow o_t \rightarrow s_{t+1},
\label{eq:interaction-transition}
\end{equation}
where $o_t$ is the resulting user response or tool observation. For a recommendation $i_t$, the environment checks whether $i_t\in\mathcal Y_x$, and the simulator generates the corresponding response. A rejection continues the interaction. The episode ends upon success, after $N_{\mathrm{rec}}$ recommendation attempts, or at horizon $T_{\max}$, yielding a trajectory $\tau=(s_1,a_1,o_1,\ldots,s_T,a_T,o_T)$.

\subsection{Tool-Augmented Recommendation}

SimRec provides three external tools through a unified interface, covering the three key information needs in recommendation: \textbf{candidate discovery}, \textbf{candidate verification}, and \textbf{user preference}. Tool outputs are incorporated into the interaction state as observations.

\noindent\textbf{Product Search.}
Given a natural-language query, the retriever returns a relevance-ranked product list. SimRec provides interchangeable retrieval backends, allowing dense or sparse retrievers to be selected for different tasks. To limit context length, search results expose compact product snippets, while full product information is accessed through the Item Detail tool.

\noindent\textbf{Item Detail.}
The Item Detail tool returns full product information for a selected candidate, providing grounded evidence for verifying constraints such as price, size, compatibility, and functionality and for distinguishing otherwise similar candidates.

\noindent\textbf{User Preference.}
For tasks with user history, $h_x$ is standardized, deduplicated, chronologically ordered with target items removed, and LLM-summarized into long-term preferences. The agent accesses this summary on demand. Preference construction details are provided in the appendix~\ref{app:preference_construction}.

Tool use is not governed by a fixed multi-step workflow. Within the prompt-defined tool budgets and action constraints, the agent decides when to retrieve, inspect items, access preferences, clarify requests, or recommend.

\subsection{Task-Anchored User Simulation}

SimRec conditions user simulation on the current task need $z_x$ rather than a reconstructed long-term user profile. The simulator generates a response to the agent's action $a_t$ as
\begin{equation}
u_t\sim\pi_{\mathrm{sim}}(\cdot\mid z_x,a_t,\mathcal H_t),
\label{eq:user-simulator}
\end{equation}
where $\mathcal H_t$ is the interaction history. For a recommendation, $a_t$ includes the proposed item $i_t$; for a clarification turn, it contains the agent's question. The simulator receives no hidden target ID and generates feedback grounded in the task need and interaction history. The environment separately determines recommendation success by checking whether $i_t\in\mathcal Y_x$. SimRec supports both a fixed external LLM as the simulator and shared-parameter self-play with role-specific prompts.

\subsection{Tool-Integrated Policy Optimization}
\subsubsection{Multi-turn GRPO}

We apply Group Relative Policy Optimization (GRPO)
\cite{shaoDeepSeekMathPushingLimits2024b} over complete multi-turn trajectories.
For each task $x$, the old policy samples a group
$\mathcal G(x)=\{\tau_1,\ldots,\tau_G\}$, where each trajectory receives
reward $R_i=R(\tau_i)$. Its group-normalized advantage is
$\hat A_i=(R_i-\mu_x)/(\sigma_x+\epsilon)$,
where $\mu_x$ and $\sigma_x$ are the mean and standard deviation of rewards
within the group. For token $y_{i,k}$ generated under context $c_{i,k}$, we define the policy
ratio as
$\rho_{i,k}(\theta)=
\pi_\theta(y_{i,k}\mid c_{i,k})/
\pi_{\theta_{\mathrm{old}}}(y_{i,k}\mid c_{i,k})$.
The per-token clipped objective is
\begin{equation}
\begin{split}
\ell_{i,k}=\min\bigl\{&\rho_{i,k}\hat A_i,\\
&\operatorname{clip}(\rho_{i,k},1-\epsilon_c,1+\epsilon_c)
\hat A_i\bigr\}.
\end{split}
\label{eq:grpo-clipped-objective}
\end{equation}
For readability, define the token-wise KL term
$D_{i,k}^{\mathrm{KL}}=D_{\mathrm{KL}}\bigl(
\pi_\theta(\cdot\mid c_{i,k})\,
\Vert\,\pi_{\mathrm{ref}}(\cdot\mid c_{i,k})\bigr)$.
The final optimization objective becomes
\begin{equation}
\begin{aligned}
J(\theta) &= \mathbb E\Bigl[\frac{1}{|\mathcal M|}
\sum_{(i,k)\in\mathcal M}\bigl(\ell_{i,k}-\beta D_{i,k}^{\mathrm{KL}}\bigr)
\Bigr].
\end{aligned}
\label{eq:grpo-objective}
\end{equation}
Here, $\mathcal M$ denotes the optimized tokens; all tokens in a
trajectory share its advantage $\hat A_i$. External-LLM simulation updates
only recommender tokens, while shared-parameter self-play jointly optimizes
both roles for recommendation success.

\subsubsection{Task-Aligned Reward}

The trajectory reward combines recommendation quality, natural retrieval, and effective tool use, with a unified penalty for redundant and overly long interactions:
\begin{equation}
\begin{aligned}
R(\tau)=\operatorname{clip}\bigl(
&\lambda_{\mathrm{rec}}R_{\mathrm{rec}}
+\lambda_{\mathrm{tool}}R_{\mathrm{tool}}\\
&+R_{\mathrm{fmt}}-P(\tau),R_{\min},R_{\max}\bigr).
\end{aligned}
\label{eq:trajectory-reward}
\end{equation}

\noindent\textbf{Recommendation Reward.}
We define $R_{\mathrm{rec}}=R_{\mathrm{hit}}+R_{\mathrm{ret}}$, where
$R_{\mathrm{hit}}=\mathbb I[i_{\mathrm{final}}\in\mathcal Y_x]$ and is zero if no final recommendation is produced. To provide a denser retrieval signal, let $\mathcal Q_{\mathrm{nr}}(\tau)$ denote non-redundant searches, i.e., previously unseen normalized queries whose Top-$K$ results contain at least one new item. We define
\begin{equation}
R_{\mathrm{ret}}(\tau)=
\max_{q_j\in\mathcal Q_{\mathrm{nr}}(\tau)}
\operatorname{NDCG}@K(L_j,\mathcal Y_x),
\label{eq:retrieval-reward}
\end{equation}
and set $R_{\mathrm{ret}}=0$ if no non-redundant search occurs. We use binary-relevance NDCG@K, computed as the ratio
of DCG@K to IDCG@K.
For single-target tasks, it reduces to the discounted gain determined by the target rank. Retrieval rewards are computed from natural, uninjected results.

\noindent\textbf{Tool-Use Reward.}
We reward valid use of each tool category:
\begin{equation}
\begin{split}
R_{\mathrm{tool}}={}&\lambda_{\mathrm{search}}R_{\mathrm{search}}
+\lambda_{\mathrm{inspect}}R_{\mathrm{inspect}}\\
&+\lambda_{\mathrm{pref}}R_{\mathrm{pref}}.
\end{split}
\label{eq:tool-reward}
\end{equation}
where $R_a=\mathbb I[N_a^{\mathrm{valid}}(\tau)>0]$ for
$a\in{\mathrm{search},\mathrm{inspect},\mathrm{pref}}$.
Invalid, failed, or redundant calls do not count, so each tool type contributes at most once per trajectory.

\noindent\textbf{Format Reward.}
$R_{\mathrm{fmt}}$ rewards each recommendation containing a parseable item ID and a nonempty description, regardless of target match.

\noindent\textbf{Interaction Penalty.}
Let $\delta_{\mathrm{rep}}(\tau)=
\mathbb I[\operatorname{norm}(q^+)=\operatorname{norm}(q^-)]$ indicate
query repetition. The interaction penalty is
\begin{equation}
\begin{aligned}
P(\tau)={}&\lambda_{\mathrm{repeat}}\delta_{\mathrm{rep}}(\tau)\\
&+\min\bigl\{P_{\max},\,\lambda_{\mathrm{len}}
[N_{\mathrm{turn}}(\tau)-T_{\mathrm{free}}]_+\bigr\}.
\end{aligned}
\label{eq:interaction-penalty}
\end{equation}
Here, $q^-$ and $q^+$ are the highest-scoring natural queries before and
after the first corrective feedback. We set $\delta_{\mathrm{rep}}=0$
if either is unavailable, and $[a]_+=\max(a,0)$.

\subsubsection{Dynamic Retrieval Mechanism}
\label{sec:retrieval-protocol}
For query $q_t$, the retriever returns $L_t=\operatorname{Retrieve}_K(q_t)$, but the agent sees only $V_t=L_t[:k]$, where $k\le K$. Reformulating queries across turns exposes different candidates and expands trajectory-level recall coverage. To address the sparse training signal over large catalogs, when a target remains outside the visible list we inject a selected $y_x\in\mathcal Y_x$ after at least $N$ user turns:
\begin{equation}
\begin{gathered}
n_{\mathrm{user}}(t)\ge N,\qquad y_x\notin V_t\\
\Longrightarrow\quad
\widetilde V_t=\operatorname{RandInsert}(V_t,y_x).
\end{gathered}
\label{eq:target-injection}
\end{equation}
Here, $\operatorname{RandInsert}$ inserts the target at a uniformly random position in the visible candidate list, avoiding a fixed positional cue. It preserves learning opportunities for candidate comparison when retrieval misses the target, following controlled candidate-construction strategies in agentic recommendation \cite{wangSelfDistilledReinforcementLearning2026a}. Retrieval rewards are always computed from uninjected results, and target injection is disabled during evaluation to prevent information leakage.

\section{Experiments}
\subsection{Experimental Setup}
\noindent\textbf{Datasets.}
We evaluate SimRec on ESCI~\cite{reddyShoppingQueriesDataset2022a}, which provides query--product relevance annotations, and Amazon C4~\cite{houBridgingLanguageItems2026a}, which contains complex shopping requests. We use five ESCI categories and four Amazon C4 categories. For Amazon C4, we collect user histories from the original Amazon Reviews 2023 data and make them available through the preference tool.

\noindent\textbf{Evaluation Metrics.} 
We report instance-weighted NDCG@100 and Recall@100 for retrieval quality, and Hit@1 for the final recommendation. For agents, retrieval metrics take the maximum across all non-redundant natural search turns on its Top-100 results, while Hit@1 measures whether the final recommended item belongs to the target set. For standalone retrievers, Hit@1 equals Recall@1. Target injection is disabled during evaluation.

\noindent\textbf{Baselines.}
We compare SimRec against these baselines: standalone retrievers (BM25~\cite{robertsonProbabilisticRelevanceFramework2009}, Qwen3 Dense~\cite{zhangQwen3EmbeddingAdvancing2025}, and BLAIR-BASE/LARGE~\cite{houBridgingLanguageItems2026a}); training-free LLM-based recommenders (GP-VisualPrompt~\cite{xiGenerativePromptingComplex2025} and ChatCRS~\cite{liChatcrsIncorporatingExternal2025}); trainable LLM-based recommenders (Rec-R1~\cite{linRecR1BridgingGenerative2026a}, RetPO~\cite{yoonAskOptimalQuestions2025}, and MiniELM~\cite{nguyenMiniELMLightweightAdaptive2025}); and LLM backbone baselines run within the SimRec framework (Qwen3-4B Base~\cite{yangQwen3TechnicalReport2025}, Qwen-Plus\footnote{\url{https://help.aliyun.com/en/model-studio/qwen-plus}}, DeepSeek-V3.2~\cite{deepseek-aiDeepSeekV32PushingFrontier2025}, DeepSeek-V4-Flash~\cite{deepseek-aiDeepSeekV4HighlyEfficient2026}, Kimi-K2.6~\cite{teamKimiK25Visual2026}, and GLM-5.1~\cite{glm-5-teamGLM5VibeCoding2026}). Agent-based methods use Qwen3 Dense by default; BM25 variants assess sparse retrieval.

\noindent\textbf{Implementation Details.}
SimRec uses Qwen3-4B with Qwen3 Dense as its default retriever. Each search returns $K=100$ items, with the top $k=20$ visible to the agent. During training, target injection is enabled after at least $N=2$ user turns if the selected target remains absent from the visible candidates. Feedback is generated by a fixed Qwen3.6-Plus simulator\footnote{\url{https://help.aliyun.com/en/model-studio/qwen3-6-plus}} by default; we evaluate shared-parameter self-play separately. GRPO uses $G=8$, a learning rate of $5\times10^{-7}$, clipping coefficient $\epsilon_c=0.2$, and KL coefficient $\beta=10^{-3}$. Agent baselines share the tool interface, search budget, candidate visibility, simulator protocol, and decoding constraints. Further details are provided in the Appendix~\ref{app:diagnostics}.

\begin{table*}[!t]
\centering
\caption{Main results on ESCI and Amazon C4. Category values are
NDCG@100; aggregate columns report NDCG@100, Recall@100, and Hit@1.
Elec. denotes Electronics. LLM-backbone baselines share SimRec's tool interface;
SimRec uses Qwen3 Dense unless labeled BM25. Bold and underline denote the
best and second-best distinct results, respectively.}
\label{tab:main-results}

\scriptsize
\setlength{\tabcolsep}{2.6pt}
\renewcommand{\arraystretch}{1.10}

\resizebox{\textwidth}{!}{%
\begin{tabular}{l*{15}{c}}
\toprule

\multirow{3}{*}{\textbf{Method}}
& \multicolumn{8}{c}{\textbf{ESCI}}
& \multicolumn{7}{c}{\textbf{Amazon C4}} \\

\cmidrule(lr){2-9}
\cmidrule(l){10-16}

& \multicolumn{5}{c}{\textbf{NDCG@100}}
& \multicolumn{3}{c}{\textbf{Avg.}}
& \multicolumn{4}{c}{\textbf{NDCG@100}}
& \multicolumn{3}{c}{\textbf{Avg.}} \\

\cmidrule(lr){2-6}
\cmidrule(lr){7-9}
\cmidrule(lr){10-13}
\cmidrule(l){14-16}

& \textbf{Baby}
& \textbf{Office}
& \textbf{Sports}
& \textbf{Game}
& \textbf{Elec.}
& \textbf{NDCG}
& \textbf{Recall}
& \textbf{Hit@1}
& \textbf{Baby}
& \textbf{Office}
& \textbf{Sports}
& \textbf{Game}
& \textbf{NDCG}
& \textbf{Recall}
& \textbf{Hit@1} \\

\midrule

\rowcolor{gray!12}
\multicolumn{16}{c}{
    \textbf{\textit{Standalone Retrieval Baselines}}
} \\

BM25
& 15.20 & 24.16 & 20.00 & 12.44 & 22.14
& 19.68 & 53.94 & 12.76
& 6.40 & 8.70 & 7.28 & 7.82
& 7.68 & 21.84 & 2.61 \\

Qwen3 Dense
& 29.06 & 35.17 & 31.74 & 34.64 & 33.07
& 32.62 & 75.69 & 19.98
& 17.34 & 18.66 & 21.77 & 18.49
& 19.80 & 52.32 & 6.45 \\

BLAIR-BASE
& 16.93 & 22.41 & 20.37 & 14.64 & 14.70
& 19.09 & 55.94 & 9.77
& 20.95 & 17.99 & 22.27 & 19.72
& 20.42 & 57.72 & 5.87 \\

BLAIR-LARGE
& 17.15 & 22.17 & 19.66 & 19.03 & 17.25
& 19.54 & 57.60 & 9.10
& 23.26 & 20.11 & 25.65 & \textbf{25.27}
& 23.42 & \textbf{63.01} & 7.14 \\

\midrule

\rowcolor{gray!12}
\multicolumn{16}{c}{
    \textbf{\textit{Training-Free LLM-based Recommenders}}
} \\

GP-VisualPrompt
& 28.31 & \textbf{36.48} & 33.47 & 35.19 & 32.99
& 33.64 & 76.03 & 22.31
& 23.14 & 19.11 & 24.17 & 18.37
& 21.82 & 56.74 & 7.14 \\

ChatCRS
& 32.46 & 34.64 & 33.60 & 38.02 & 35.15
& 34.36 & 76.80 & 22.31
& 21.87 & 20.08 & 24.65 & 17.82
& 22.12 & 57.96 & 7.61 \\

\midrule

\rowcolor{gray!12}
\multicolumn{16}{c}{
    \textbf{\textit{SimRec: LLM Backbone Baselines}}
} \\

Qwen3-4B Base
& 28.61 & 33.62 & 31.93 & 35.42 & 31.36
& 32.23 & 72.03 & 21.20
& 21.01 & 20.59 & 25.80 & 20.89
& 22.94 & 56.21 & 8.30 \\

Qwen-Plus
& 29.32 & 32.40 & 30.42 & 36.27 & 33.61
& 31.73 & 70.92 & 33.52
& 17.79 & 16.31 & 20.89 & 16.16
& 18.50 & 48.61 & 12.66 \\

DeepSeek-V3.2
& 22.45 & 31.13 & 32.02 & 34.64 & 31.61
& 31.00 & 70.81 & 28.52
& \underline{23.93} & \textbf{23.03}
& 26.71 & \underline{21.39}
& \underline{24.62} & 58.36 & 13.41 \\

DeepSeek-V4-Flash
& 29.94 & 35.15 & 34.21 & 34.65 & 33.53
& 33.89 & 73.14 & 33.96
& 20.71 & 19.46 & 24.28 & 18.40
& 21.64 & 54.88 & 12.78 \\

Kimi-K2.6
& 30.39 & 35.60 & 33.71
& \underline{38.53} & \underline{35.75}
& 34.50 & 77.25 & \textbf{40.40}
& 22.26 & 20.28 & 25.55 & 18.45
& 22.69 & 56.80 & \textbf{16.26} \\

GLM-5.1
& \underline{32.93} & 35.56 & \textbf{34.42}
& 38.37 & 34.95
& \underline{34.99} & \underline{77.80} & \underline{36.96}
& 23.73 & 22.32 & \underline{27.43} & 20.80
& 24.61 & \underline{59.52} & \underline{16.09} \\

\midrule

\rowcolor{gray!12}
\multicolumn{16}{c}{
    \textbf{\textit{Trainable LLM-based Recommenders}}
} \\

Rec-R1
& 27.39 & 34.39 & 30.98 & 35.00 & 33.46
& 32.01 & 57.20 & 20.75
& 19.34 & 16.55 & 19.13 & 16.22
& 18.04 & 50.55 & 4.59 \\

RetPO (DPO)
& 23.16 & 28.28 & 25.91 & 31.27 & 34.29
& 27.65 & 61.26 & 22.64
& 20.85 & 18.85 & 23.30 & 19.39
& 21.12 & 56.04 & 6.40 \\

MiniELM (SFT+DPO)
& 28.66 & 34.57 & 32.02 & 35.85 & 33.92
& 32.81 & 74.70 & 21.42
& 22.29 & 18.62 & 22.59 & 19.21
& 20.92 & 54.35 & 6.97 \\

\addlinespace[2pt]

\rowcolor{blue!6}
\textbf{SimRec-4B}
& \textbf{33.28} & \underline{35.62} & 34.11
& \textbf{38.80} & \textbf{39.02}
& \textbf{35.42} & \textbf{77.91} & 24.08
& \textbf{24.23} & \underline{22.81}
& \textbf{27.50} & 19.03
& \textbf{24.72} & 59.18 & 9.76 \\

\rowcolor{blue!6}
\hspace{0.8em}+\;Self-Play
& 32.46 & 35.39 & \underline{34.39}
& 37.59 & 34.23
& 34.72 & 75.81 & 22.42
& 17.88 & 18.83 & 21.69 & 16.90
& 19.77 & 49.54 & 7.14 \\

\rowcolor{blue!6}
\hspace{0.8em}+\;BM25
& 21.06 & 25.04 & 23.81 & 21.54 & 28.51
& 24.02 & 65.37 & 14.76
& 15.32 & 14.96 & 16.36 & 17.91
& 15.88 & 40.77 & 5.75 \\

\rowcolor{blue!6}
\hspace{0.8em}+\;Self-Play + BM25
& 24.46 & 29.72 & 29.80 & 22.22 & 26.11
& 27.88 & 58.27 & 13.65
& 15.24 & 15.09 & 15.81 & 15.93
& 15.50 & 39.43 & 6.04 \\

\bottomrule

\end{tabular}%
}

\end{table*}

\subsection{Main Results}

\noindent\textbf{Overall Performance.}
On ESCI (Table~\ref{tab:main-results}), SimRec-4B improves NDCG@100, Recall@100, and Hit@1 from 32.23/72.03/21.20 to 35.42/77.91/24.08, respectively, outperforming compared training methods. NDCG@100 improves over Qwen3-4B Base in all five product categories, most strongly in Electronics (+7.66 percentage points). SimRec also exceeds the strongest alternative backbone's NDCG (GLM-5.1, 34.99) and the task-aligned baselines. These results support the introduction's distinction between
supplying an agent with tools and learning how to use them
over a complete interaction; the behavioral analyses below
examine what changes in the policy.

\noindent\textbf{Retrieval coverage and final choice remain distinct bottlenecks.} On Amazon C4, SimRec reaches the best aggregate NDCG@100 of 24.72 versus 22.94 for Qwen3-4B Base, Recall@100 and Hit@1 increase by 2.97 and 1.45 percentage points respectively. However, BLAIR-LARGE, an Amazon-trained retriever, achieves higher Recall (63.01 vs.\ 59.18), while larger LLMs such as Kimi-K2.6 achieve higher Hit@1. SimRec nevertheless achieves the highest aggregate NDCG@100 among the compared methods. This suggests that its main advantage lies in better candidate retrieval rather than retrieval recall alone.

\noindent\textbf{Different Retrieval Backends.}
To test robustness to the retriever, we replace Qwen3 Dense with BM25. SimRec-4B improves NDCG@100/Recall@100 from 19.68/53.94 to 24.02/65.37 on ESCI and from 7.68/21.84 to 15.88/40.77 on Amazon C4 over standalone BM25, showing that its gains persist with a weaker sparse backend. The larger relative gains further suggest that interaction can partly compensate for weaker retrieval. Retrieval gains also extend to the 1.7B backbone (Appendix~\ref{app:small-model}); the evaluated general benchmarks show no systematic degradation after post-training (Appendix~\ref{app:general-capabilities}).

\subsection{Ablation Study}
\label{sec:ablation}

\begin{table}[!t]
\centering
\centering
\captionof{table}{Component ablation on Amazon C4. Drops in parentheses are percentage points.}
\label{tab:ablation-amazon}
\small
\setlength{\tabcolsep}{2pt}
\renewcommand{\arraystretch}{1.07}
\begin{tabularx}{\linewidth}{
  >{\raggedright\arraybackslash}X
  r@{\hspace{2pt}}l
  @{\hspace{8pt}}
  r@{\hspace{2pt}}l
}
\toprule
\rowcolor{black!6}
\textbf{Variant}
& \multicolumn{2}{c}{\textbf{NDCG@100}}
& \multicolumn{2}{c}{\textbf{Hit@1}} \\
\midrule

\rowcolor{blue!6}
\textbf{Full SimRec}
& \textbf{24.72} & 
& \textbf{9.76} & \\

w/o Preference
& 21.17 & \abldrop{3.55}
& 7.65 & \abldrop{2.11} \\

w/o Item Detail
& 23.42 & \abldrop{1.30}
& 6.04 & \abldrop{3.72} \\

w/o Feedback
& 22.14 & \abldrop{2.58}
& 8.06 & \abldrop{1.70} \\

\bottomrule
\end{tabularx}
\vspace{5pt}
\centering
\captionof{table}{Reward ablation on ESCI.}
\label{tab:reward-ablation-esci}
\small
\setlength{\tabcolsep}{2pt}
\renewcommand{\arraystretch}{1.07}
\begin{tabularx}{\linewidth}{
  >{\raggedright\arraybackslash}X
  r@{\hspace{2pt}}l
  @{\hspace{8pt}}
  r@{\hspace{2pt}}l
}
\toprule
\rowcolor{black!6}
\textbf{Reward Setting}
& \multicolumn{2}{c}{\textbf{NDCG@100}}
& \multicolumn{2}{c}{\textbf{Hit@1}} \\
\midrule

\rowcolor{blue!6}
\textbf{Full Reward}
& \textbf{32.38} &
& \textbf{17.22} & \\

w/o Hit@1
& 25.45 & \abldrop{6.93}
& 14.28 & \abldrop{2.94} \\

w/o Retrieval
& 26.18 & \abldrop{6.20}
& 14.58 & \abldrop{2.64} \\

w/o Tool
& 25.49 & \abldrop{6.89}
& 14.34 & \abldrop{2.88} \\

\bottomrule
\end{tabularx}
\end{table}

\noindent\textbf{Different components address distinct recommendation needs.}
We conduct component ablations on Amazon C4 (Table~\ref{tab:ablation-amazon}).
Removing user preferences causes the largest NDCG@100 drop (3.55 points),
whereas removing item-detail inspection yields the largest Hit@1 drop(3.72 points). This distinction suggests that preferences help guide candidate retrieval and ranking, while item details support more detailed
verification before recommendation decision. Disabling multi-turn feedback further
reduces NDCG@100 and Hit@1, respectively. Thus, even with preference and item-detail tools available, the agent cannot fully compensate for the absence of corrective user feedback,
which provides task-specific information for revising earlier decisions.
Further domain-level results are provided in
Appendix~\ref{app:additional-ablation}.

\noindent\textbf{Intermediate rewards and the final hit serve complementary purposes.}
At training step 500 on ESCI (Table~\ref{tab:reward-ablation-esci}),
removing retrieval or tool-use rewards reduces Hit@1 by 2.64 or 2.88 points, despite retaining the final-hit objective. These results suggest that rewarding evidence acquisition helps improve downstream recommendation decisions rather than merely
encouraging tool use. Conversely, removing the hit reward lowers NDCG@100 by 6.93 points and Hit@1 by 2.94 points, indicating that
intermediate rewards alone cannot replace the final outcome signal. Together, these findings support jointly optimizing evidence
acquisition and recommendation accuracy through trajectory-level rewards.

\subsection{Analysis}
\label{sec:analysis}
We examine whether the aggregate gains reflect the interaction behaviors highlighted in the introduction: useful feedback, recovery after rejection, tool use, and candidate ranking.
\subsubsection{User Simulation Modes}

\begin{table}[!htbp]
\centering
\caption{Feedback-mode comparison on ESCI.}
\label{tab:simulator-feedback-comparison}

\small
\setlength{\tabcolsep}{2pt}
\renewcommand{\arraystretch}{1.08}
\begin{tabular}{@{}lccc@{}}
\toprule
\rowcolor{gray!12}
\textbf{Feedback Mode}
& \shortstack{\textbf{NDCG@10}\\(\%) $\uparrow$}
& \shortstack{\textbf{Hit@1}\\(\%) $\uparrow$}
& \shortstack{\textbf{Invalid}\\(\%) $\downarrow$} \\
\midrule
Fixed Feedback$^{\dagger}$ & \underline{31.52} & 18.65 & \textbf{11.90} \\
Qwen3-4B Base & 31.10 & \underline{19.53} & 30.70 \\
\rowcolor{blue!6}
\textbf{SimRec-4B (Self-Play)} & \textbf{32.15} & \textbf{20.64} & \underline{14.00} \\
\bottomrule
\end{tabular}

\par\vspace{2pt}
{\footnotesize\raggedright
$^{\dagger}$ Fixed Feedback always returns:
``This recommendation does not meet my needs.''
Invalid rate is the percentage of malformed outputs or invalid tool calls.\par}

\end{table}

\noindent\textbf{Self-play performance depends on the interaction setting.}
With dense retrieval, self-play approaches external simulation
on ESCI (34.72 vs.\ 35.42 NDCG@100) but trails it on Amazon C4
(19.77 vs.\ 24.72). With BM25, it leads on ESCI
(27.88 vs.\ 24.02) and nearly matches on Amazon C4
(15.50 vs.\ 15.88). Since simulator stability, model capacity,
and parameter sharing differ, this comparison cannot isolate
their effects. Self-play avoids an external simulator but
does not consistently match it (Appendix~\ref{app:additional-ablation}).

\noindent\textbf{Self-play balances recommendation quality and stability.}
On ESCI (Table~\ref{tab:simulator-feedback-comparison}),
Fixed Feedback has fewer invalid outputs (11.90\% vs.\ 14.00\%),
but self-play achieves higher NDCG@10 (32.15 vs.\ 31.52)
and Hit@1 (20.64 vs.\ 18.65). Self-play also reduces
the base policy's invalid rate from 30.70\% to 14.00\%.
Its shorter median feedback (114 vs.\ 195 characters)
does not establish a causal effect of length.

\noindent\textbf{Simulator choice affects recommendation performance.} We evaluate SimRec with GLM-5.1 and Kimi-K2.6 in place of the default Qwen3.6-Plus simulator, keeping the remaining setup fixed. GLM-5.1 improves NDCG, Recall and Hit@1 to 36.93\%, 79.02\%, and 26.42\%, respectively. Kimi-K2.6 slightly improves Recall@100 but lowers NDCG@100 and Hit@1 (Appendix~\ref{app:cross-simulator}). These results highlight the impact of simulator choice on downstream recommendation performance.

\subsubsection{Recovery Analysis}
\label{sec:recovery}
\begin{figure}[!htbp]
    \centering
    \includegraphics[width=0.96\columnwidth]{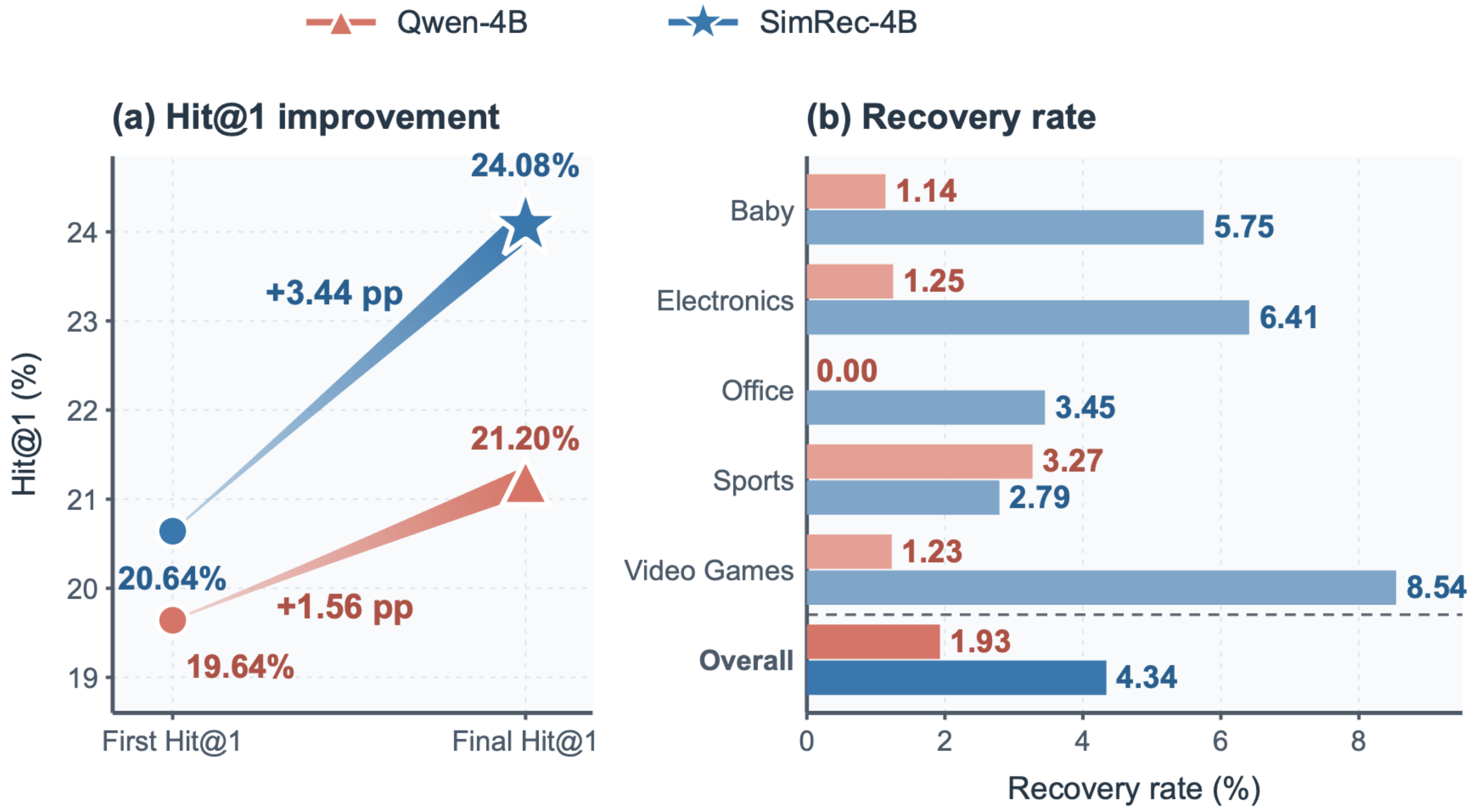}
    \caption{Recovery analysis on ESCI. (a) Hit@1 improvement from the first to the final recommendation for Qwen3-4B Base and SimRec-4B. (b) Category-wise and overall recovery rates among initial misses.}
    \label{fig:recovery_analysis}
\end{figure}

\begin{figure*}[!t]
\centering
\includegraphics[width=0.92\textwidth]{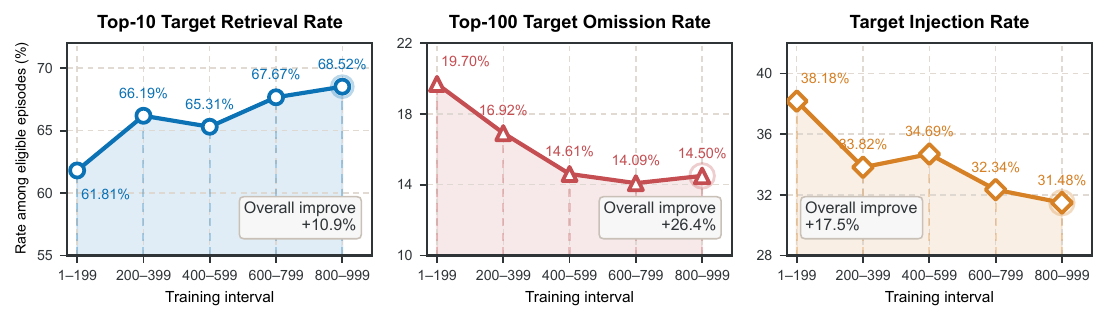}
\caption{Training dynamics of natural retrieval and target injection. As training progresses, the Top-10 hit rate increases and the
Top-100 miss rate decreases, indicating improved retrieval quality.}
\label{fig:augmentation-dynamics}
\end{figure*}

\noindent\textbf{SimRec improves recovery from initial misses.}
Recovery is measured as the fraction of unsuccessful first recommendations
that become hits by the final turn. On ESCI, Hit@1 rises from 19.64\%
to 21.20\% for Qwen3-4B Base and from 20.64\% to 24.08\% for
SimRec-4B (Figure~\ref{fig:recovery_analysis}).
SimRec achieves a larger first-to-final improvement(+3.44 vs.\ +1.56 percentage points), with an overall recovery
rate of 4.34\% compared with 1.93\% for the base model,
more than doubling the recovery rate.
Category-level results show higher recovery rates in four of
five categories, with particularly large gains in Video Games
and Electronics, while Sports shows a slight decline.
These results suggest that interaction training improves the
agent's ability to refine recommendations following corrective
feedback, although recovery remains challenging across categories.

\subsubsection{Controlled Reranking}
\label{sec:reranking}
\begin{table}[!htbp]
\centering
\caption{Controlled ESCI reranking (NDCG@10, \%).
Base denotes Qwen3-4B; Self-Play denotes SimRec-4B with self-play.
Game and Elec. denote Video Games and Electronics.}
\label{tab:esci-reranking}
\footnotesize
\setlength{\tabcolsep}{2.0pt}
\renewcommand{\arraystretch}{1.10}
\begin{tabular}{@{}lrrrrrr@{}}
\toprule
\rowcolor{gray!12}
\textbf{Model} & \textbf{Baby} & \textbf{Office} & \textbf{Sport}
& \textbf{Game} & \textbf{Elec.} & \textbf{Avg.} \\
\midrule
Shuffled & 20.91 & 26.05 & 18.39 & 14.49 & 13.56 & 19.73 \\
Base & 29.27 & 27.33 & 23.51 & 23.57 & 21.08 & 24.75 \\
Self-Play & 33.63 & 28.18 & 22.99 & \textbf{27.63} & 21.97 & 25.57 \\
\rowcolor{blue!8}
\textbf{SimRec-4B} & \textbf{35.06} & \textbf{31.48}
& \textbf{25.54} & 24.07 & \textbf{27.21} & \textbf{28.06} \\
\bottomrule
\end{tabular}
\end{table}
\textbf{Interaction training also improves reranking ability.} To disentangle retrieval recall from candidate reranking ability, we construct a controlled generative reranking task on ESCI. For each query, all models receive the same 16 candidate items and rerank them from a randomized input order, while retrieval scores and source information are hidden. SimRec-4B improves overall NDCG@10 from 24.75 to \textbf{28.06} over Qwen3-4B Base, while self-play reaches 25.57 (Table~\ref{tab:esci-reranking}). With the fixed candidate set, this gain indicates SimRec improves candidate comparison and product-level ranking based on query and item evidence through training. Dataset construction details are provided in Appendix~\ref{app:reranking-details}.

\subsubsection{Training Dynamics}
\label{sec:augmentation-dynamics}

\textbf{Natural retrieval improves while target injection declines.} From the first to the last training window (Figure~\ref{fig:augmentation-dynamics}), the Top-10 hit rate rises from 61.81\% to 68.52\%, indicating that targets are increasingly ranked near the top. Meanwhile, the Top-100 miss rate falls from 19.70\% to 14.50\%, indicating improved retrieval coverage. Early in training, 38.18\% of eligible trajectories require target injection; even after multiple turns, the target is still absent from the candidates visible to the agent. The injection rate subsequently declines to 31.48\% as natural retrieval improves. Detailed statistics are provided in Appendix~\ref{app:additional_training_diagnostics}. 

\textbf{SimRec learns a more active tool-use policy.} SimRec also exhibits a more active evidence-seeking policy, with average tool use increasing from 2.43 to 5.19 calls per session. Most of this increase comes from item-detail inspection (0.58 to 1.83 calls) and preference-tool calls (0.02 to 0.86), indicating greater use of both product- and user-specific evidence. Moreover, sessions with more calls have lower average task reward (Appendix~\ref{app:tool_behavior}), consistent with harder sessions prompting further investigation. Together with the component ablations, these results suggest that training changes how SimRec gathers evidence (Section~\ref{sec:ablation}) during interaction.

\section{Conclusion}
We present \textbf{SimRec}, which trains a tool-integrated recommendation agent over complete interactions using task-anchored feedback and trajectory-level rewards. SimRec outperforms retrieval, training-free, and task-aligned baselines in overall NDCG@100 on ESCI and Amazon C4, while improving over its untrained 4B backbone. Fixed-candidate reranking and recovery analyses further show better product selection and more effective revision after rejection. These results demonstrate the value of learning to gather evidence, recommend, and revise within a single interaction policy.

\section*{Limitations}
SimRec relies on LLM-based user simulation to train multi-turn recommendation policies. Although feedback is grounded in the current task need, simulated responses may not fully capture real users' preferences and behaviors. Improving simulator fidelity and evaluating SimRec through real-user interactions remain important directions for future work.

To address sparse recommendation rewards, SimRec employs target injection during training. Although injection is disabled during evaluation, this introduces a discrepancy between training and inference. Developing more effective training strategies without target injection could further improve the robustness of the learned policy.

Our experiments primarily cover ESCI and Amazon C4 with Qwen3-4B and a smaller 1.7B backbone. Performance also varies across user simulators and retrieval backends. Generalization to larger models, broader product domains, and more diverse interaction settings remains to be explored.

\FloatBarrier
\bibliography{custom}

\clearpage
\appendix

\section{Additional Experimental Details}
\label{app:diagnostics}
\subsection{Implementation}
The recommender is initialized from Qwen3-4B. The retrieval backend returns $K=100$ items, with the top $k=20$ visible by default. Training-only target injection is enabled after $N=2$ user turns if the target remains absent from the visible list. GRPO uses learning rate $5\times10^{-7}$, rollout batch size 8, group size $G=8$, temperature 1.0, top-p 1.0, clipping coefficient 0.2, and KL coefficient $10^{-3}$. The external simulator is Qwen3.6-Plus, with temperature 0.2, maximum output length 220 tokens and seed 42. Rewards are clipped to $[0,2.5]$. Duplicate queries receive no additional valid-search bonus; unchanged feedback queries incur a 0.05 penalty. Length penalties start after four turns, increase by 0.02 per additional turn and are capped at 0.20. The recommendation reward weight and retrieval NDCG scale are both 1.0. The format reward is 0.01; valid search, item-detail, and preference bonuses are 0.03, 0.01, and 0.01.

\subsection{All Results}
Tables~\ref{tab:esci-main} and~\ref{tab:amazon-c4-main}
report category-level NDCG@100 and Recall@100, together with
aggregate NDCG@100, Recall@100, and Hit@1.
\begin{table*}[t]
\centering
\caption{Detailed ESCI results. Category columns report NDCG@100
and Recall@100; AVG additionally reports Hit@1. SimRec uses Qwen3 Dense
unless marked BM25. Bold and underline indicate best and second-best
distinct values, respectively (including ties).}
\label{tab:esci-main}

\scriptsize
\setlength{\tabcolsep}{3.0pt}
\renewcommand{\arraystretch}{1.10}

\resizebox{\textwidth}{!}{%
\begin{tabular}{lccccccccccccc}
\toprule
\textbf{Method}
& \multicolumn{2}{c}{\textbf{Baby}}
& \multicolumn{2}{c}{\textbf{Office}}
& \multicolumn{2}{c}{\textbf{Sports}}
& \multicolumn{2}{c}{\textbf{Video Games}}
& \multicolumn{2}{c}{\textbf{Electronics}}
& \multicolumn{3}{c}{\textbf{AVG}} \\
\cmidrule(lr){2-3}
\cmidrule(lr){4-5}
\cmidrule(lr){6-7}
\cmidrule(lr){8-9}
\cmidrule(lr){10-11}
\cmidrule(l){12-14}
& \textbf{NDCG} & \textbf{Recall}
& \textbf{NDCG} & \textbf{Recall}
& \textbf{NDCG} & \textbf{Recall}
& \textbf{NDCG} & \textbf{Recall}
& \textbf{NDCG} & \textbf{Recall}
& \textbf{NDCG} & \textbf{Recall} & \textbf{Hit@1} \\
\midrule

\rowcolor{gray!12}
\multicolumn{14}{c}{\textbf{\textit{Retrieval Baselines}}} \\
BM25
& 15.20 & 45.63 & 24.16 & 63.04 & 20.00 & 53.92 & 12.44 & 41.75 & 22.14 & 58.25
& 19.68 & 53.94 & 12.76 \\

Qwen3 Dense
& 29.06 & 70.87 & 35.17 & \underline{78.80} & 31.74 & 75.25 & 34.64 & 78.64 & 33.07 & 73.79
& 32.62 & 75.69 & 19.98 \\

BLAIR-BASE
& 16.93 & 54.37 & 22.41 & 63.59 & 20.37 & 58.09 & 14.64 & 41.75 & 14.70 & 49.51
& 19.09 & 55.94 & 9.77 \\

BLAIR-LARGE
& 17.15 & 54.37 & 22.17 & 63.04 & 19.66 & 58.82 & 19.03 & 51.46 & 17.25 & 52.43
& 19.54 & 57.60 & 9.10 \\

\midrule
\rowcolor{gray!12}
\multicolumn{14}{c}{\textbf{\textit{Training-Free LLM-based Recommenders}}} \\

GP-VisualPrompt
& 28.31 & 70.87 & \textbf{36.48} & \textbf{79.35} & 33.47 & 76.23 & 35.19 & 71.84 & 32.99 & 78.64
& 33.64 & 76.03 & 22.31 \\

ChatCRS
& 32.46 & \underline{75.73} & 34.64 & 73.37 & 33.60 & \underline{76.96} & 38.02 & \underline{80.58} & 35.15 & \underline{79.61}
& 34.36 & 76.80 & 22.31 \\

\midrule
\rowcolor{gray!12}
\multicolumn{14}{c}{\textbf{\textit{SimRec: LLM Backbone Baselines}}} \\

Qwen3-4B Base
& 28.61 & 67.96 & 33.62 & 72.28 & 31.93 & 72.55 & 35.42 & 74.76 & 31.36 & 70.87
& 32.23 & 72.03 & 21.20 \\

Qwen-Plus
& 29.32 & 68.93 & 32.40 & 70.65 & 30.42 & 69.36 & 36.27 & 75.73 & 33.61 & 74.76
& 31.73 & 70.92 & 33.52 \\

DeepSeek-V3.2
& 22.45 & 60.19 & 31.13 & 71.74 & 32.02 & 70.83 & 34.64 & 77.67 & 31.61 & 72.82
& 31.00 & 70.81 & 28.52 \\

DeepSeek-V4-Flash
& 29.94 & 69.90 & 35.15 & 72.28 & 34.21 & 72.79 & 34.65 & 73.79 & 33.53 & 78.64
& 33.89 & 73.14 & 33.96 \\

Kimi-K2.6
& 30.39 & 72.82 & 35.60 & 77.17 & 33.71 & \textbf{77.45} & \underline{38.53} & 79.61 & \underline{35.75} & 78.64
& 34.50 & 77.25 & \textbf{40.40} \\

GLM-5.1
& \underline{32.93} & 73.79 & 35.56 & \underline{78.80} & \textbf{34.42} & \textbf{77.45} & 38.37 & 79.61 & 34.95 & \underline{79.61}
& \underline{34.99} & \underline{77.80} & \underline{36.96} \\

\midrule
\rowcolor{gray!12}
\multicolumn{14}{c}{\textbf{\textit{Trainable LLM-based Recommenders}}} \\

Rec-R1
& 27.39 & 52.10 & 34.39 & 58.74 & 30.98 & 55.43 & 35.00 & 62.59 & 33.46 & 61.19
& 32.01 & 57.20 & 20.75 \\

RetPO (DPO)
& 23.16 & 52.43 & 28.28 & 58.70 & 25.91 & 59.31 & 31.27 & 66.99 & 34.29 & 76.70
& 27.65 & 61.26 & 22.64 \\

MiniELM (SFT+DPO)
& 28.66 & 69.90 & 34.57 & 77.72 & 32.02 & 74.51 & 35.85 & 74.76 & 33.92 & 74.76
& 32.81 & 74.70 & 21.42 \\

\rowcolor{blue!6}
\textbf{SimRec-4B}
& \textbf{33.28} & \textbf{78.64} & \underline{35.62} & 78.26 & 34.11 & 75.49 & \textbf{38.80} & \textbf{81.55} & \textbf{39.02} & \textbf{82.52}
& \textbf{35.42} & \textbf{77.91} & 24.08 \\

\rowcolor{blue!6}
\hspace{0.8em}+\;Self-Play
& 32.46 & 72.82 & 35.39 & 76.09 & \underline{34.39} & 76.72 & 37.59 & 74.76 & 34.23 & 75.73
& 34.72 & 75.81 & 22.42 \\

\rowcolor{blue!6}
\hspace{0.8em}+\;BM25
& 21.06 & 61.17 & 25.04 & 69.57 & 23.81 & 65.69 & 21.54 & 55.34 & 28.51 & 70.87
& 24.02 & 65.37 & 14.76 \\

\rowcolor{blue!6}
\hspace{0.8em}+\;Self-Play + BM25
& 24.46 & 57.28 & 29.72 & 61.96 & 29.80 & 61.03 & 22.22 & 46.60 & 26.11 & 53.40
& 27.88 & 58.27 & 13.65 \\

\bottomrule
\end{tabular}%
}

\end{table*}

\begin{table*}[t]
\centering
\caption{Detailed Amazon C4 results. Category columns report
NDCG@100 and Recall@100; AVG additionally reports Hit@1.
SimRec uses Qwen3 Dense unless marked BM25.
Bold and underline indicate best and second-best values, respectively.}
\label{tab:amazon-c4-main}

\scriptsize
\setlength{\tabcolsep}{3.2pt}
\renewcommand{\arraystretch}{1.10}

\resizebox{\textwidth}{!}{%
\begin{tabular}{lccccccccccc}
\toprule
\textbf{Method}
& \multicolumn{2}{c}{\textbf{Baby}}
& \multicolumn{2}{c}{\textbf{Office}}
& \multicolumn{2}{c}{\textbf{Sports}}
& \multicolumn{2}{c}{\textbf{Video Games}}
& \multicolumn{3}{c}{\textbf{AVG}} \\
\cmidrule(lr){2-3}
\cmidrule(lr){4-5}
\cmidrule(lr){6-7}
\cmidrule(lr){8-9}
\cmidrule(l){10-12}
& \textbf{NDCG} & \textbf{Recall}
& \textbf{NDCG} & \textbf{Recall}
& \textbf{NDCG} & \textbf{Recall}
& \textbf{NDCG} & \textbf{Recall}
& \textbf{NDCG} & \textbf{Recall} & \textbf{Hit@1} \\
\midrule
\rowcolor{gray!12}\multicolumn{12}{c}{\textbf{\textit{Retrieval Baselines}}} \\
BM25
& 6.40 & 21.05 & 8.70 & 24.26 & 7.28 & 20.45 & 7.82 & 20.67 & 7.68 & 21.84 & 2.61 \\
Qwen3 Dense
& 17.34 & 51.82 & 18.66 & 48.53 & 21.77 & 55.61 & 18.49 & \underline{51.33} & 19.80 & 52.32 & 6.45 \\
BLAIR-BASE
& 20.95 & \underline{61.94} & 17.99 & 52.17 & 22.27 & 62.03 & 19.72 & 50.67 & 20.42 & 57.72 & 5.87 \\
BLAIR-LARGE
& 23.26 & \textbf{65.18} & 20.11 & \textbf{58.75} & 25.65 & \textbf{65.11} & \textbf{25.27} & \textbf{65.33} & 23.42 & \textbf{63.01} & 7.14 \\

\midrule
\rowcolor{gray!12}\multicolumn{12}{c}{\textbf{\textit{Training-Free LLM-based Recommenders}}} \\
GP-VisualPrompt
& 23.14 & 60.32 & 19.11 & 53.03 & 24.17 & 59.49 & 18.37 & \underline{51.33} & 21.82 & 56.74 & 7.14 \\
ChatCRS
& 21.87 & 56.28 & 20.08 & 55.98 & 24.65 & 62.03 & 17.82 & 48.00 & 22.12 & 57.96 & 7.61 \\

\midrule
\rowcolor{gray!12}\multicolumn{12}{c}{\textbf{\textit{SimRec: LLM Backbone Baselines}}} \\
Qwen3-4B Base
& 21.01 & 55.87 & 20.59 & 52.86 & 25.80 & 59.89 & 20.89 & \underline{51.33} & 22.94 & 56.21 & 8.30 \\
Qwen-Plus
& 17.79 & 48.99 & 16.31 & 46.79 & 20.89 & 51.34 & 16.16 & 41.33 & 18.50 & 48.61 & 12.66 \\
DeepSeek-V3.2
& \underline{23.93} & 57.89 & \textbf{23.03} & 56.67 & 26.71 & 61.76 & \underline{21.39} & 48.67 & \underline{24.62} & 58.36 & 13.41 \\
DeepSeek-V4-Flash
& 20.71 & 57.09 & 19.46 & 51.82 & 24.28 & 58.69 & 18.40 & 44.00 & 21.64 & 54.88 & 12.78 \\
Kimi-K2.6
& 22.26 & 55.47 & 20.28 & 54.77 & 25.55 & 60.70 & 18.45 & 47.33 & 22.69 & 56.80 & \textbf{16.26} \\
GLM-5.1
& 23.73 & 58.70 & 22.32 & \underline{58.41} & \underline{27.43} & 62.83 & 20.80 & 48.67 & 24.61 & \underline{59.52} & \underline{16.09} \\

\midrule
\rowcolor{gray!12}\multicolumn{12}{c}{\textbf{\textit{Trainable LLM-based Recommenders}}} \\
Rec-R1
& 19.34 & 56.74 & 16.55 & 49.38 & 19.13 & 50.43 & 16.22 & 45.47 & 18.04 & 50.55 & 4.59 \\
RetPO (DPO)
& 20.85 & 58.70 & 18.85 & 52.69 & 23.30 & 59.09 & 19.39 & 49.33 & 21.12 & 56.04 & 6.40 \\
MiniELM (SFT+DPO)
& 22.29 & 56.68 & 18.62 & 51.99 & 22.59 & 56.15 & 19.21 & 50.67 & 20.92 & 54.35 & 6.97 \\

\rowcolor{blue!6}
\textbf{SimRec-4B}
& \textbf{24.23} & 58.70 & \underline{22.81} & 56.67 & \textbf{27.50} & \underline{63.10} & 19.03 & 50.00 & \textbf{24.72} & 59.18 & 9.76 \\
\rowcolor{blue!6}
\hspace{0.8em}+\;Self-Play
& 17.88 & 46.56 & 18.83 & 49.57 & 21.69 & 51.87 & 16.90 & 42.67 & 19.77 & 49.54 & 7.14 \\
\rowcolor{blue!6}
\hspace{0.8em}+\;BM25
& 15.32 & 43.32 & 14.96 & 39.17 & 16.36 & 40.91 & 17.91 & 42.00 & 15.88 & 40.77 & 5.75 \\
\rowcolor{blue!6}
\hspace{0.8em}+\;Self-Play + BM25
& 15.24 & 41.70 & 15.09 & 38.99 & 15.81 & 38.90 & 15.93 & 40.00 & 15.50 & 39.43 & 6.04 \\
\bottomrule
\end{tabular}%
}

\end{table*}

\subsection{Controlled Reranking Construction}
\label{app:reranking-details}
We construct the ESCI reranking set with BM25 as the only upstream retriever. BM25 searches the product \texttt{contents} field with $k_1=1.2$ and $b=0.75$, returns the Top-20 candidates, and filters out instances whose upstream NDCG@20 is zero. We retain the first 16 candidates after filtering, resulting in 345 valid examples from the 901-instance local ESCI test set across Baby Products, Office Products, Sports and Outdoors, Video Games, and Electronics. To prevent the model from copying the upstream order, each candidate list is shuffled with random seed 42, while \texttt{retrieval\_score} and \texttt{retrieval\_type} are hidden. The random input order serves as the no-reranking baseline.

\subsection{Additional 1.7B Results}
\label{app:small-model}

We additionally evaluate SimRec with a 1.7B Qwen backbone on ESCI. Table~\ref{tab:small-model} reports results under dense and BM25 retrieval for both external simulation and shared-parameter self-play. These runs extend the evaluation to a smaller model scale. Turns denotes the average number of interaction turns and is reported as a descriptive efficiency statistic. \textit{--} indicates that none of the 901 samples produces a natural recommendation, leaving no valid sample for Hit@1 computation.

\begin{table}[!htbp]
\centering
\caption{ESCI results with Qwen3-1.7B. Dense refers to Qwen3 Dense;
metrics are percentages and Turns is the average number of turns.}
\label{tab:small-model}
\footnotesize
\setlength{\tabcolsep}{3pt}
\renewcommand{\arraystretch}{1.12}
\begin{tabular}{@{}lrrrr@{}}
\toprule
\rowcolor{gray!12}
\textbf{Setting} & \textbf{NDCG} & \textbf{Recall} & \textbf{Hit@1} & \textbf{Turns} \\
\midrule
Base + Dense         & 33.47 & 75.92 & -- & 8.23 \\
Base + BM25          & 19.63 & 53.50 & -- & 7.49 \\
SimRec + Dense       & 34.17 & 77.03 & 21.31 & 4.79 \\
SimRec + BM25        & 29.91 & 71.81 & 18.20 & 3.84 \\
Self-Play + Dense    & 34.42 & 76.25 & 21.53 & 4.00 \\
Self-Play + BM25     & 25.76 & 65.48 & 17.20 & 4.02 \\
\bottomrule
\end{tabular}
\end{table}

\subsection{General Benchmark Retention}
\label{app:general-capabilities}
Recommendation-oriented training retains performance across the evaluated general benchmarks, with higher scores on four of five. Performance improves on MMLU, GSM8K, IFEval, and MBPP, with HumanEval remaining within 0.61 percentage points of the base score.

\begin{table}[!htbp]
\centering
\caption{General-capability performance before and after recommendation training. Values are percentages; $\Delta$ denotes RL$-$Base in percentage points.}
\label{tab:general-capabilities}

\small
\setlength{\tabcolsep}{7pt}
\renewcommand{\arraystretch}{1.18}

\begin{tabular}{@{}lrrr@{}}
\toprule
\rowcolor{gray!12}
\textbf{Benchmark} & \textbf{Base} & \textbf{RL} & \textbf{$\Delta$} \\
\midrule
MMLU & 55.83 & \textbf{57.93} & $+2.10$ \\
GSM8K & 78.92 & \textbf{82.49} & $+3.57$ \\
IFEval & 79.11 & \textbf{81.70} & $+2.59$ \\
MBPP pass@1 & 44.00 & \textbf{52.00} & $+8.00$ \\
HumanEval pass@1 & \textbf{71.95} & 71.34 & $-0.61$ \\
\bottomrule
\end{tabular}

\end{table}
\subsection{Domain-level Ablations and Tool Behavior}
Across the four Amazon C4 categories, removing user preferences or multi-turn feedback generally reduces NDCG@100 and Recall@100, suggesting that both provide useful signals for interactive recommendation. Removing item-detail inspection increases Recall@100 from 59.18 to 63.65, but reduces NDCG@100 from 24.72 to 23.42 and Hit@1 from 9.76 to 6.04. This suggests that broader retrieval coverage does not necessarily translate into better final recommendations, highlighting the value of item-level verification. (Table~\ref{tab:amazon-ablation-main}).

\begin{table*}[t]
\centering
\caption{Amazon C4 component ablations across four categories. Values are percentages reported as NDCG@100 / Recall@100.}
\label{tab:amazon-ablation-main}

\small
\setlength{\tabcolsep}{4pt}
\renewcommand{\arraystretch}{1.12}

\begin{tabular}{@{}lccccc@{}}
\toprule
\rowcolor{gray!12}
\textbf{Ablation} & \textbf{Baby} & \textbf{Office} & \textbf{Sports} & \textbf{Video Games} & \textbf{ALL} \\
\midrule
w/o \texttt{get\_user\_preference} & 20.72 / 55.06 & 19.09 / 51.65 & 23.30 / 56.82 & 19.34 / 46.00 & 21.17 / 53.89 \\
w/o \texttt{get\_item\_details} & \textbf{24.88 / 69.64} & \textbf{21.24 / 60.49} & \textbf{25.13 / 64.71} & \textbf{20.93 / 60.67} & \textbf{23.42 / 63.65} \\
w/o multi-turn feedback & 20.03 / 54.66 & 20.24 / 54.42 & 24.56 / 57.62 & 20.86 / 50.00 & 22.14 / 55.46 \\
\bottomrule
\end{tabular}

\end{table*}

\begin{table*}[t]
\centering
\caption{Behavioral diagnostics for the Amazon C4 component ablations. These metrics characterize behavioral differences accompanying the main NDCG@100 and Recall@100 results.}
\label{tab:amazon-ablation-behavior}

\small
\setlength{\tabcolsep}{3pt}
\renewcommand{\arraystretch}{1.12}

\begin{tabular}{@{}lrrrrrrr@{}}
\toprule
\rowcolor{gray!12}
\textbf{Ablation} & \textbf{Turns} & \textbf{Search} & \textbf{Details} & \textbf{Preference} & \textbf{Eff. pref.} & \textbf{Feedback} & \textbf{Len. pen.} \\
\midrule
w/o \texttt{get\_user\_preference} & 4.99 & 99.88\% & 4.53\% & 0\% & 0\% & 0.26\% & 0.0213 \\
w/o \texttt{get\_item\_details} & 6.71 & 100\% & 0\% & 100\% & 19.11\% & 0.58\% & 0.0542 \\
w/o multi-turn feedback & 3.99 & 99.77\% & 99.25\% & 76.77\% & 2.33\% & 0\% & 0 \\
\bottomrule
\end{tabular}

\end{table*}

\section{Additional Ablation Analyses}
\label{app:additional-ablation}

\subsection{Self-Play Optimization Analysis}

\begin{table}[!htbp]
\centering
\caption{Self-play optimization variants at step 500.}
\label{tab:selfplay-optimization}

\scriptsize
\setlength{\tabcolsep}{4pt}
\renewcommand{\arraystretch}{1.05}
\begin{tabular}{lcc}
\toprule
\rowcolor{gray!12}
\textbf{Setting}
& \textbf{NDCG@100}
& \textbf{Hit@1} \\
\midrule
Train user, no reward sharing
& \textbf{21.75}
& \textbf{9.38} \\
Train user, reward sharing
& 20.97
& 7.03 \\
No user-token training
& 20.57
& 7.81 \\
\bottomrule
\end{tabular}

\end{table}

We examine the effects of optimizing user-role tokens
and sharing trajectory rewards during self-play.
As shown in Table~\ref{tab:selfplay-optimization},
disabling user-token training yields the lowest NDCG@100.
Among the evaluated variants, training user-role tokens
without explicit reward sharing achieves the highest
NDCG@100 (21.75) and Hit@1 (9.38).
Explicit reward sharing does not improve either metric
at step 500.

\subsection{Cross-Simulator Evaluation}
\label{app:cross-simulator}

To examine how the choice of user simulator affects SimRec, we conduct a
cross-simulator evaluation using Qwen3.6-Plus, Kimi-K2.6, and GLM-5.1 as
user simulators while keeping the remaining experimental configuration
unchanged. Table~\ref{tab:cross-simulator} reports the average performance
across categories.

\begin{table}[!htbp]
\centering
\caption{Cross-simulator evaluation of SimRec. Changes ($\Delta$)
are percentage points relative to Qwen3.6-Plus.}
\label{tab:cross-simulator}
\footnotesize
\setlength{\tabcolsep}{4pt}
\renewcommand{\arraystretch}{1.10}
\begin{tabular}{@{}lrrr@{}}
\toprule
\rowcolor{gray!12}
\textbf{Metric} & \textbf{Qwen3.6-Plus} & \textbf{Kimi-K2.6} & \textbf{GLM-5.1} \\
\midrule
NDCG@100 & \underline{35.42} & 34.41 & \textbf{36.93} \\
\quad $\Delta$ (pp) & -- & $-1.01$ & $\mathbf{+1.51}$ \\
Recall@100 & 77.91 & \underline{78.69} & \textbf{79.02} \\
\quad $\Delta$ (pp) & -- & $+0.78$ & $\mathbf{+1.11}$ \\
Hit@1 & \underline{24.08} & 22.42 & \textbf{26.42} \\
\quad $\Delta$ (pp) & -- & $-1.66$ & $\mathbf{+2.34}$ \\
\bottomrule
\end{tabular}
\end{table}
The results show a clear sensitivity to the choice of user simulator.
Relative to Qwen3.6-Plus, Kimi-K2.6 increases Recall@100 by
$0.78$ percentage points, but decreases NDCG@100 by $1.01$ points
and Hit@1 by $1.66$ points. In contrast, GLM-5.1 consistently
improves all three metrics, yielding gains of $1.51$, $1.11$, and $2.34$ percentage points in NDCG@100, Recall@100,
and Hit@1, respectively.

These results indicate that the user simulator is not merely an interchangeable
component: its quality can meaningfully affect the recommendation policy learned
or evaluated through interaction. These results suggest that simulator choice can substantially affect downstream recommendation performance, although they do not directly establish differences in feedback fidelity or quality. This cross-simulator comparison therefore further
supports the role of the user simulator as an important component of SimRec,
while also showing that the framework is compatible with multiple simulator
backbones.

\subsection{Additional Training Diagnostics}
\label{app:additional_training_diagnostics}
\begin{table*}[t]
\centering
\caption{Training-stage dynamics of target injection. Eligible episodes satisfy the minimum user-turn condition. The injection rate additionally requires that the target be absent from the model-visible top $k=20$.}
\label{tab:augmentation-dynamic}

\small
\setlength{\tabcolsep}{5pt}
\renewcommand{\arraystretch}{1.12}

\begin{tabular}{@{}lrrrrrrr@{}}
\toprule
\rowcolor{gray!12}
\textbf{Training steps} & \textbf{Episodes} & \textbf{Eligible} & \textbf{Top-10 hit} & \textbf{Rank 11--100} & \textbf{Top-100 miss} & \textbf{Injection} & \textbf{Inj./eligible} \\
\midrule
1--199 & 6,368 & 64.62\% & 61.81\% & 18.49\% & 19.70\% & 24.67\% & 38.18\% \\
200--399 & 6,400 & 71.47\% & 66.19\% & 16.90\% & 16.92\% & 24.17\% & 33.82\% \\
400--599 & 6,400 & 72.92\% & 65.31\% & 20.08\% & 14.61\% & 25.30\% & 34.69\% \\
600--799 & 6,400 & 70.30\% & 67.67\% & 18.24\% & 14.09\% & 22.73\% & 32.34\% \\
800--999 & 6,240 & 72.18\% & 68.52\% & 16.98\% & 14.50\% & 22.72\% & 31.48\% \\
\bottomrule
\end{tabular}

\end{table*}
Across 994 rollout files, there are 31,808 episodes and 54,352 searches; 7,610 episodes contain target injection. Comparing the first and last 50 rollout files, natural Top-10 hit rises from 56.29\% to 65.20\%, natural Top-100 hit rises from 75.14\% to 84.45\%, and episode-level injection falls from 28.62\% to 24.75\%. These windows provide a complementary view of the stage-level training dynamics.

\subsection{Tool-Use Behavior Analysis}
\label{app:tool_behavior}

We further analyze the tool-use behavior of SimRec on the test sessions. We compare against
the Qwen3-4B base policy evaluated on the same test set.

\paragraph{Tool-use frequency.}
Table~\ref{tab:tool_usage_statistics} summarizes the average number of tool
calls and interaction turns per session. SimRec invokes external tools more
actively than the base policy, with 5.19 tool calls per session compared with
2.43 for Qwen3-4B. The largest increases come from item-detail inspection
(1.83 vs.\ 0.58 calls per session) and preference queries
(0.86 vs.\ 0.02), while the increase in search calls is more moderate
(2.50 vs.\ 1.83). SimRec also produces longer interaction trajectories on
average (9.05 vs.\ 6.28 turns).

Importantly, the tool-use reward in SimRec is not assigned independently to
every tool invocation. Instead, it serves as a trajectory-level binary
shaping signal for each tool type: once a tool type has been invoked validly
within a session, subsequent calls to the same tool type do not provide
additional tool-use reward. Multiple calls remain permissible when they are
needed for task completion---for example, the policy may inspect several
different candidate items using the item-detail tool---but these additional
calls do not accumulate extra $R_{\mathrm{tool}}$. The purpose of this reward
component is therefore to encourage the policy to learn valid use of the
available tools, rather than to reward a larger number of calls.

\begin{table}[!htbp]
    \centering
\caption{
    Tool-use statistics on the Amazon session test set.
    SimRec uses external tools more actively than the Qwen3-4B base policy,
    especially item-detail and preference calls; Base denotes Qwen3-4B Base.
    }
\label{tab:tool_usage_statistics}
    \small
    \setlength{\tabcolsep}{4pt}
\begin{tabular}{@{}lcc@{}}
        \toprule
        \rowcolor{gray!12}
        \textbf{Metric} & \textbf{SimRec} & \textbf{Base} \\
        \midrule
        Tool calls / session       & 5.19 & 2.43 \\
        Search calls / session     & 2.50 & 1.83 \\
        Item-detail calls / session& 1.83 & 0.58 \\
        Preference calls / session & 0.86 & 0.02 \\
        Turns / session            & 9.05 & 6.28 \\
        \bottomrule
    \end{tabular}

\end{table}

\paragraph{Tool-call count and task performance.}
To examine whether the higher tool-use frequency could itself provide a
systematic reward advantage, we stratify SimRec trajectories by their total
number of tool calls. The results are shown in
Table~\ref{tab:tool_count_performance}. Recommendation quality and final
trajectory reward do not increase with the number of tool calls. Sessions
with zero to two calls achieve the highest NDCG@100, Recall@100, Hit@1, and
average reward, whereas sessions involving five or more calls are
substantially more difficult.

\begin{table}[!htbp]
\centering
\caption{SimRec performance by tool-call count on the 1,722-session
Amazon C4 test set. Quality metrics use a 0--1 scale; group differences
are observational, not causal.}
\label{tab:tool_count_performance}
\small
\setlength{\tabcolsep}{5pt}
\renewcommand{\arraystretch}{1.10}
\begin{tabular}{@{}lrrr@{}}
\toprule
\rowcolor{gray!12}
\textbf{Metric} & \textbf{0--2} & \textbf{3--4} & \textbf{$\geq 5$} \\
\midrule
Sessions & 120 & 268 & 1,334 \\
NDCG@100 & 0.9949 & 0.5274 & 0.1237 \\
Recall@100 & 1.0000 & 0.9925 & 0.4745 \\
Hit@1 & 0.9917 & 0.1381 & 0.0090 \\
Avg. reward & 2.2383 & 0.8719 & 0.3528 \\
\bottomrule
\end{tabular}
\end{table}
The strong association between longer tool-use trajectories and lower task
performance should not be interpreted as evidence that additional tool use
causes worse recommendations. A more plausible explanation is selection by
task difficulty: easy sessions can often be resolved after the initial
retrieval or a small number of inspections, while difficult or unresolved
sessions trigger further search, item inspection, and preference
consultation. Nevertheless, this analysis rules out a simple interpretation
in which repeatedly invoking tools provides a reliable way to increase the
trajectory-level objective. In particular, repeated calls to an already used
tool type cannot accumulate additional $R_{\mathrm{tool}}$, and empirically
the final reward does not increase with total tool-call count.

Overall, the analysis suggests that post-training changes not only the final
ranking behavior but also the interaction policy: SimRec is more willing to
consult external evidence, especially item details and user preferences, when
the current recommendation remains unresolved. At the same time,
$R_{\mathrm{tool}}$ should be understood as a lightweight tool-validity
shaping signal rather than a direct measure of the marginal utility of each
individual tool invocation.

\section{Prompts Used in the Recommendation Framework}
\label{app:prompts}
This appendix reports the prompts used for the recommendation agent, user simulator, product reranking, and user-preference-profile construction. Text enclosed in braces (e.g., \texttt{\{reference\_query\}}) denotes a runtime placeholder that is replaced with the corresponding instance-specific value. The recommendation and auxiliary reranking tasks use distinct output protocols; the prompts below are reproduced verbatim.

\subsection{Recommendation Agent}
\begin{PromptBlock}{Recommendation Agent Prompt} 
You are the recommender in a shopping dialogue.
Your goal is to find the single best catalog item for the shopper.
Use retrieval evidence before recommending. Do not guess product facts.
Keep each tool decision concise and reserve enough space for the final structured recommendation.

Valid output modes:
1. Tool-call mode
- Use this mode when you need search results or item details.
- Output exactly one JSON tool call and nothing else.
- Do not add explanation, markdown, role labels, or natural language around the tool call.
- search format: {"name":"search_products","arguments":{"query":"..."}}
- item-details format: {"name":"get_item_details","arguments":{"item_id":"B0XXXXXXXX"}}
- user-preference format: {"name":"get_user_preference","arguments":{}}

2. Shopper-facing reply mode
- Use this mode only after you have enough retrieved evidence to answer.
- Write plain English, including the required field labels below.
- Never expose tool names, JSON, code, commands, role tokens, XML tags, or hidden reasoning.
- Never end with an unfinished sentence, ellipsis, or placeholder.

3. Clarification/dialogue mode
- Use this mode only when the shopper's request lacks a decisive constraint and no reasonable search query can be formed.
- Ask exactly one short question and nothing else.
- Do not emit a tool call, recommendation fields, JSON, or internal reasoning in this mode.
- After the shopper answers, return to tool-call mode and search with the clarified constraint.

Recommendation output contract:
- Recommend exactly one item unless the shopper explicitly asks for alternatives.
- Every recommendation must use exactly this field structure:
  recommended_item_id: <ITEM_ID>
  recommended_item_description: <2-4 complete English sentences>
- The description is the most important field. It must explain why this item matches the shopper's requested attributes.
- The description should briefly explain why the recommended item
  matches the shopper's stated constraints and preferences.
- Mention only user-relevant attributes supported by retrieved evidence.
- Do not add unsupported attributes or keyword lists.
- Mention only claims supported by retrieved evidence.
- Any shopper-facing answer after tool use must still use the recommendation field structure; do not answer follow-up questions as free-form prose.

Decision and tool policy:
- Build search queries from the current request's product type and discriminative constraints. Keep them concise; use at most {N} searches total.
- Use get_item_details selectively to verify plausible candidates and hard constraints; do not inspect the same item twice.
- Treat get_user_preference as optional ranking evidence, never as a required first step. Use it at most once and only when one missing soft preference could change a specific search term or break a tie among relevant candidates.
- Useful cases are broad or taste-driven requests, unresolved soft trade-offs, or several similarly relevant candidates. Skip it when explicit attributes already determine the search or when search/item details can resolve the uncertainty.
- When top search results are similar, or shopper feedback adds specific attributes such as brand, model, size, edition, material, compatibility, or included accessories, inspect one plausible candidate with get_item_details before recommending.
- Recommend only when retrieved evidence supports the shopper's requested product type and attributes.
- After shopper feedback, do not answer in prose; search with the corrected constraints, inspect useful evidence if needed, then recommend using the required fields.
- Never repeat a previous search query. If no meaningfully different query is possible or the tool budget is nearly exhausted, recommend the best grounded retrieved item.
- Ask one short clarification question only when no reasonable search query can be formed.
\end{PromptBlock}

\subsection{User Simulator Prompt}

\subsubsection{User Simulator: Initial Shopping Request}

\begin{PromptBlock}{Initial Shopping Request Prompt}
You are simulating the shopper in a recommendation dialogue. State your shopping need clearly in natural English. Keep it concise, realistic, and specific enough for a recommender to search. Never output, quote, or reveal any product or item ID, even if it appears in the context.
Reference query: {reference_query}
\end{PromptBlock}

\subsubsection{User Simulator: Conversational Feedback}
\begin{PromptBlock}{Conversational Feedback Prompt}
You are simulating the shopper in a recommendation dialogue. Stay in character as a shopper with a clear product need. Continue the conversation naturally in plain English. If the assistant asks a question, answer it directly. Give corrective feedback with one or two concrete missing requirements so the recommender can search again. Do not use rigid labels or templates; sound like a normal shopper. Do not output JSON, role labels, tags, hidden metadata, or any product/item ID. Never reveal, quote, or repeat an Item ID, even if it appears in the context.
\end{PromptBlock}

\subsubsection{User Input Template}

\begin{PromptBlock}{User Input Template}
Your shopping need is based on:
- Reference query: {reference_query}
Assistant recommended item Ids: {recommended_item_ids}
Recommendation was grounded in recent tool results: {grounded}

Recent dialogue:
{dialogue}

Latest assistant message:
{assistant_text}

Write the next shopper reply only.
\end{PromptBlock}

\subsection{Listwise Product Reranking}
\subsubsection{System Prompt}
\begin{PromptBlock}{Listwise Product Reranking}
You are a helpful AI assistant. You first think about the reasoning process in the mind and then provide the user with the answer.
\end{PromptBlock}

\subsubsection{User Input Template}
\begin{PromptBlock}{User Input Template}
You are an expert in product reranking. Given a customer query and a list of candidate products, rerank the products so the most relevant items are placed at the top. Consider semantic meaning, product attributes, and customer intent.

Below are the candidate products (item_id and metadata):
```json
{candidate_block}
```

Below is the query:
```text
{query}
```

Show your work in <think> tags. Your final response must be in JSON format within <answer> </answer> tags. The JSON must contain the reranked list of item_ids in order of relevance under the key "reranked_items". The length of the reranked list should match the number of input candidates.
\end{PromptBlock}

\subsection{User Preference Profile Construction}
\label{app:preference_construction}
\subsubsection{Prompt Template}
 We strictly remove the target item of the current task from the user history before preference summarization, preventing the preference profile from directly exposing the ground-truth recommendation.

\begin{PromptBlock}{User Preference Profile Construction Prompt}
Based on the user's purchase history below, generate a concise user preference profile.

## User Purchase History (chronological order):
{purchase_history}

## Statistics:
- Average rating: {avg_rating}

## Requirements:
1. Summarize the user's likely purchase preferences and favorite product types.
2. If low-rated items clearly indicate dislikes, mention that briefly.
3. Keep the profile to exactly one short natural sentence of 20--30 words.
4. Output ONLY in the <answer> tag, no other explanatory text.
5. Do not include unsupported details, user IDs, bullet points, or multiple sentences.

## Output Format:
<answer>
The user likely prefers practical, durable products in their favorite categories and avoids items with clear quality or usability issues.
</answer>
\end{PromptBlock}
\end{document}